\documentclass[sigconf,table]{acmart}

\usepackage{graphicx}
\usepackage{tikz}
\usepackage{caption}
\usepackage{subcaption}
\usepackage{listings}
\usepackage{algpseudocode}
\usepackage{hyperref}
\usepackage[normalem]{ulem}
\usepackage[nodisplayskipstretch]{setspace}

\usetikzlibrary{arrows}

\useunder{\uline}{\ul}{}

\copyrightyear{2022} 
\acmYear{2022} 
\setcopyright{acmlicensed}\acmConference[GECCO '22]{Genetic and Evolutionary Computation Conference}{July 9--13, 2022}{Boston, MA, USA}
\acmBooktitle{Genetic and Evolutionary Computation Conference (GECCO '22), July 9--13, 2022, Boston, MA, USA}
\acmPrice{15.00}
\acmDOI{10.1145/3512290.3528866}
\acmISBN{978-1-4503-9237-2/22/07}

\begin{document}
\title{Functional Code Building Genetic Programming}

\author{Edward Pantridge}
\orcid{0000-0003-0535-5268}
\affiliation{
  \institution{Swoop}
  \city{Cambridge} 
  \state{Massachusetts} 
  \country{USA}
}
\email{ed@swoop.com}

\author{Thomas Helmuth}
\orcid{0000-0002-2330-6809}
\affiliation{
	\institution{Hamilton College}
	\city{Clinton}
	\state{New York}
	\country{USA}
}
\email{thelmuth@hamilton.edu}

\author{Lee Spector}
\orcid{0000-0001-5299-4797}
\affiliation{
	\institution{Amherst College, Hampshire College, and UMass Amherst}
	\city{Amherst}
	\state{Massachusetts}
	\country{USA}
}
\email{lspector@amherst.edu}

\begin{abstract}
General program synthesis has become an important application area for genetic programming (GP), and for artificial intelligence more generally. Code Building Genetic Programming (CBGP) is a recently introduced GP method for general program synthesis that leverages reflection and first class specifications to support the evolution of programs that may use arbitrary data types, polymorphism, and functions drawn from existing codebases. However, neither a formal description nor a thorough benchmarking of CBGP have yet been reported. In this work, we formalize the method of CBGP using algorithms from type theory. Specially, we show that a functional programming language and a Hindley-Milner type system can be used to evolve type-safe programs using the process abstractly described in the original CBGP paper. Furthermore, we perform a comprehensive analysis of the search performance of this functional variant of CBGP compared to other contemporary GP program synthesis methods.
\end{abstract}

%
%



\maketitle


\section{Introduction}
\label{sec:intro}

A variety of genetic programming (GP) methods have been developed for general program synthesis applications, in which the goal is to automatically produce programs of the kind that humans routinely write~\cite{Sobania:2021:RecentDevInProgSynthWithEvo}. Such programs usually make use of a range of data and control structures.

Among the most successful recent approaches are those using a multi-type stack-based language called Push~\cite{Spector:2005:push3, Helmuth:2015:BenchmarkSuite} and those rooted in the use of grammars, such as Grammatical Evolution (GE)~\cite{Hemberg:2019:DomainKnowledgeAndNoveltyImproveGE} and Grammar Guided Genetic Programming (G3P)~\cite{Forstenlechner:2017:G3P, Forstenlechner:2018:G3P-extention}.

In the Push approach, evolved programs are expressed in a novel language that was designed to allow for the use of arbitrary data and control structures without imposing constraints on program syntax. This is accomplished by arranging for the passing of arguments between instructions to take place via typed data stacks. The result is that Push programs (or, more commonly, linear ``plushy'' genomes that map in a straightforward way to Push programs) can be freely mutated and crossed over without risking syntax or execution errors, even when multiple types, conditionals, loops, and more sophisticated data and control structures are being used~\cite{Pantridge:2019:GPTP}.

PushGP systems---GP systems that evolve Push programs---have produced some of the best general program synthesis results to date, but the interactions between Push programs, human programming practices, and human-written codebases leave something to be desired. For example, although it is possible to ``wrap'' arbitrary functions as Push instructions that can then be used in evolved Push programs, this is a cumbersome, manual process that becomes impossible with the introduction of polymorphic types and functions. If one wants to use an evolved Push program in software that is written in a conventional programming language, then one either has to translate the evolved Push program into the conventional language or to include a Push interpreter in the software. Because is difficult for humans to read, understand, debug, or modify Push programs, translation is not always straightforward, and inclusion of evolved Push programs in production software may raise quality and reliability concerns.

By contrast, grammar-based approaches support general program synthesis by embracing the syntactic specification of data and control structures, and by employing methods that always generate syntactically valid programs in a conventional programming language. This provides a better interface to human programming practices and codebases. However, polymorphism is still problematic because grammars do not fully capture type information, which limits the possibly of real-world applications. In addition, these methods do not yet appear to be able to solve benchmark software synthesis problems quite as reliably as PushGP.

Code building genetic programming (CBGP) is a recently developed method that takes some inspiration from both of these approaches. It uses linear genomes that are similar to PushGP's plushy genomes, and which are compiled in a manner similar to Push program execution~\cite{Pantridge:2019:GPTP}. With CBGP, however, this compilation process produces code in a conventional programming language, which can be read and understood by humans, and which can also make use of arbitrary (and possibly polymorphic) functions. As noted in the original CBGP paper~\cite{Pantridge:2020:GECCO:CBGP}, this appears to provide benefits both in terms of problem-solving power and potential for interaction with human programming practices.

The original CBGP paper, however, was a preliminary report, which provided neither a formal description of the algorithm nor a thorough benchmarking of CBGP relative to other methods. In this paper we fill those gaps. We formalize CBGP using a functional, Hindley-Milner type system, and we describe our new functional CBGP system on the basis of this formalization. We then perform a comprehensive analysis of the search performance of this system, comparing it to other contemporary GP program synthesis methods.

In the next section we review the results demonstrated in the original CBGP paper and flesh out the goals for the present work. In the subsequent section we present tools from type theory that we will use in our formalization. We then present functional CBGP in detail. This is followed by a section on the design of our experiments, and then by a section on our experimental results. We conclude with a discussion of the implications of our results and of avenues for future work.

\section{Code Building Genetic Programming}

CBGP is a method of evolving programs using a stack-based compiler which folds variable length, linear, genomes into type-safe abstract syntax trees (AST). The introduction of CBGP in 2020 included demonstrations of novel capabilities such as:
\begin{itemize}
    \item Type safe programs that call polymorphic functions.
    \item Anonymous functions defined inside evolved programs.
    \item Higher order functions for control flow.
    \item Use of user-defined data types, such as classes.
    \item Dynamic generation of the function set (aka genetic source~\cite{Helmuth:2020:ALife:source}).
\end{itemize}
In addition, the evolved programs are generated as source code in a conventional programming language, a feature which was previously only possible using grammar based GP methods. These features were demonstrated on a small set of simple benchmark problems~\cite{Pantridge:2020:GECCO:CBGP}.

This paper focuses on a subset of these capabilities and aims to clarify and standardize the implementation by leveraging properties of functional programming and type theory. In particular, the functional CBGP system described in this paper uses parametric polymorphism, anonymous function creation, and higher order functions. We only consider a finite set of data types common to other GP program synthesis systems such as PushGP, GE, and G3P. Also, the function set used in this version of CBGP was manually implemented and curated to be as similar as possible to the function sets found in other contemporary GP program synthesis systems.

The ``host language'' of a program synthesis system is the language in which synthesized programs are expressed. The version of CBGP proposed in this paper requires a functional programming language, and Clojure was selected because the LISP syntax makes it trivial to convert ASTs into source code and vice versa. The choice of host language is an implementation decision and the algorithms detailed in this paper are applicable to any functional host language.

\section{Tools from Type Theory}
\label{sec:HM}

Since the invention of simply typed lambda calculus by Church in 1940, the field of type theory has been developing algorithms for analyzing the validity of functional programs based solely on their structure~\cite{Church:1940:SimplyTypedLambdaCalculus}. One relatively modern manifestation of these ideas is the Hindley-Milner (HM) type system, which is commonly used by compilers of modern functional programming languages~\cite{Hindley:1969:HMTypeSystem, Milner:1978:HMTypeSystem}. It analyses programs represented as abstract syntax trees (AST) to prove the program is type safe. Furthermore, the HM system can provide the (possibly polymorphic) data type returned by the AST. We briefly describe the kinds of AST expressions the HM type system can analyze in the following paragraphs.

The atomic expressions, or leaves of the ASTs, are literals (aka constants) and variables. A variable is a symbol that will be translated into a known value when the program is executed. It is assumed the data type of a literal is trivially known and the data type of a variable is given by a mapping called the ``type environment'' denoted as $\Gamma$.

Larger expression trees can be created by composing literals and variables with function abstraction, function application, and \texttt{let} binding expressions. A function abstraction expression denotes the creation of a function with some fresh (new and unique) variables as arguments and an expression tree representing the body. A function application expression denotes the calling of a function on some arguments, each given by a child expression. A \texttt{let} binding expression defines a local variable which will take the value of a given child expression. 

\begin{figure*}
    \centering
    \begin{subfigure}[b]{0.3\textwidth}
        \textbf{Python}
        \begin{lstlisting}
f = lambda x: max(0, x)
map(f, input)
        \end{lstlisting} 
        
        \textbf{Haskell}
        \begin{lstlisting}
let f = \x -> max 0 x 
in map f input
        \end{lstlisting} 
        
        \textbf{LISP (Clojure)}
        \begin{lstlisting}
(let [f (fn [x] (max 0 x))]
  (map f input))
        \end{lstlisting} 
        \caption{The source code in a variety of languages.}
        \label{fig:neg-to-zero-code}
    \end{subfigure}
    \hfill
    \begin{subfigure}[b]{0.3\textwidth}
        \centering
        \begin{tikzpicture}
            \node (1) {LET};
            \node (3) [below left of=1] {ABS};
            \node (2) [left of=3] {f};
            \node (4) [below left of=3] {x};
            \node (5) [right of=4] {APP};
            \node (6) [below left of=5] {max};
            \node (7) [below of=5] {0 : Int};
            \node (8) [below right of=5] {x};
            \node (9) [below right of=1] {APP};
            \node (10) [right of=5] {map};
            \node (11) [right of=10] {f};
            \node (12) [right of=11] {input};
            \path[every node]
                (1) edge node [] {} (2)
                (1) edge node [] {} (3)
                (1) edge node [] {} (9)
                (3) edge node [] {} (4)
                (3) edge node [] {} (5)
                (5) edge node [] {} (6)
                (5) edge node [] {} (7)
                (5) edge node [] {} (8)
                (9) edge node [] {} (10)
                (9) edge node [] {} (11)
                (9) edge node [] {} (12);
        \end{tikzpicture}
        \caption{The abstract syntax tree of the program. \texttt{ABS} denotes function abstraction and \texttt{APP} denotes function application. Literals are tagged with their data type.}
        \label{fig:neg-to-zero-ast}
    \end{subfigure}
    \hfill
    \begin{subfigure}[b]{0.35\textwidth}
        \centering
        \begin{tabular}{l r}
            \toprule
            \textbf{Var} & \textbf{Type} \\
            \midrule
            input & Seq[Int] \\
            max & (Int, Int) $\rightarrow$ Int \\
            map & $\forall \alpha, \beta . ((\alpha \rightarrow \beta), Seq[\alpha]) \rightarrow Seq[\beta]$ \\
            \bottomrule
        \end{tabular}
        \caption{The type environment used to prove the type safety of the AST. Notice that the types of local variables, $x$ and $f$, are not required because they can be inferred.}
        \label{fig:neg-to-zero-env}
    \end{subfigure}
    \caption{An example program which solves the negative-to-zero benchmark problem. The HM type system is able to prove that this program is type safe and will return a $Seq[Int]$ given the variable types provided by the environment in Figure \ref{fig:neg-to-zero-env}.}
    \label{fig:neg-to-zero}
\end{figure*}

Figure \ref{fig:neg-to-zero-code} and \ref{fig:neg-to-zero-ast} show the source code and AST of an example program that the HM type system is capable of analysing.

\subsection{Types}
\label{sec:types}

The primitive, atomic, types in the HM system are called ``ground'' types. The set of ground types typically includes: $Boolean$, $Int$, $Double$, $Char$, and $String$ although there is some variation between implementations. In our work, we consider $Nil$ a ground type.

A type constructor is an operator that builds a type from one or more other types. Collection types such as $Sequence$, $Set$, and $Map$ are type constructors that must be provided the data type of their elements to build the full type. For example, we denote a set of strings as $Set[String]$. The function type constructor takes one or more argument types and a return type. For example, the function type of the string length function is $(String) \rightarrow Int$.

The Hindley-Milner type system supports abstract types using a form of parametric polymorphism called type ``schemes''~\cite{Hindley:1969:HMTypeSystem, Damas:1982:TypeScheme}. The canonical example of a type scheme is the \texttt{identity} function. The argument type of the \texttt{identity} function can be any type, but the return type will always match the argument type. The type of the \texttt{identity} function is $\forall \alpha . \alpha \rightarrow \alpha$ which is read as ``for all possible types $\alpha$, the function type from $\alpha$ to $\alpha$''.

A more complex example is the polymorphic type of the \texttt{map} higher order function. 
\[
\forall \alpha, \beta . ((\alpha \rightarrow \beta), Sequence[\alpha]) \rightarrow Sequence[\beta]
\]
We read this as: ``For all possible types $\alpha$ and $\beta$, a function which takes a function from $\alpha$ to $\beta$ and a sequence of $\alpha$ and returns a sequence of $\beta$.'' The parameters of a scheme are referred to as ``type variables''. In the examples above, $\alpha$ and $\beta$ are type variables.

\subsection{Unification}
\label{sec:unification}

The Hindley-Milner type system is often used by compilers and code analyzers to perform type checking and type inference of abstract syntax trees. A crucial component of these algorithms is \textit{unification}; a procedure which produces a set of type substitutions that will bind free type variables to concrete types, if possible. Unification has roots in theorem proving and solving systems of symbolic equations~\cite{Robinson:1965:Unification, Martelli:1982:Unification}.

To illustrate the use of unification, suppose we have a function, $f$, with the type $\forall \alpha . (Sequence[\alpha]) \rightarrow \alpha$, and an expression, $e$, of type $Sequence[Int]$. To determine if a type safe expression can be created by applying $f$ to $e$, we unify the argument type of $f$ with the type of $e$. The unification algorithm will succeed and provide the substitution ${\alpha := Int}$. Using this, we can replace all instances of $\alpha$ with $Int$ in the type of the new composite expression to determine that $f(e)$ has type $Int$.

If the unification of two types requires a type variable be bound to more than one concrete type, or if unification requires two different non-variable types to be the same, for example $Int = Boolean$, the AST fails type checking.

\section{Functional Code Building GP}

In this section, we describe how functional CBGP uses HM types, schemes, and unification to produce type safe programs during evolution.\footnote{The source code for our implementation is available at https://github.com/erp12/cbgp-lite.}

\subsection{Genomes}
\label{sec:genomes}

\begin{figure}
    \centering
    \texttt{
    [OPEN, LocalVar(1), Literal(0), Var(max), APP, CLOSE, ABS[Int], OPEN, Var(input), LocalVar(1), Var(map), APP, CLOSE, LET]
    }
    \caption{One possible genome which compiles to the AST shown Figure \ref{fig:neg-to-zero-ast} under the type environment in Figure \ref{fig:neg-to-zero-env}.}
    \label{fig:neg-to-zero-genome}
\end{figure}

CBGP uses a variable length linear genome, which is translated into a type-safe AST using a compilation process. Specifically, the plushy genome structure commonly used in PushGP systems is used in CBGP~\cite{Pantridge:2019:GPTP}. The categories of genes that can be in a genome sequence are: literals, variables, local variables, stack instructions, and structure tokens.

Literal and variable expressions (as described in Section \ref{sec:HM}), as well as local variable genes, correspond to the production of leaves in the resulting ASTs. Local variable genes contain a baked-in index (natural number) value that will resolve to a variable during the compilation process, as discussed below.

The stack instruction genes affect the construction of the internal nodes of the compiled ASTs. In contrast to PushGP, only a small set of stack instructions are required to implement CBGP. Each stack instruction corresponds one of the non-leaf expression types of ASTs described in Section \ref{sec:HM}. Specifically, the full set of stack instruction gene types is: \texttt{APP}, \texttt{ABS}, and \texttt{LET}. The \texttt{APP} gene produces a new subtree for function application, and the \texttt{LET} gene produces a new subtree for local variable binding. Function abstraction genes are annotated with the data type of the function's arguments. For example, the \texttt{ABS[Int]} gene will attempt to compile an AST that creates a single-argument function with the argument type of $Int$. This is a distinct gene from \texttt{ABS[Int,String]}. The \texttt{ABS[]} gene denotes a zero-argument function.

The "structure tokens" of plushy genomes can be either an \texttt{open} or \texttt{close} gene. These denote the start and end of nested "chunks" of genome, which are leveraged by the compilation process discussed in Section \ref{sec:compilation}. Using the plushy translation algorithm, the linear genomes are transformed into nested sequences such that slices of genome between corresponding \texttt{open} and \texttt{close} tokens are included in a nested sequence. This translation process is robust to unbalanced numbers of \texttt{open} and \texttt{close} genes~\cite{Pantridge:2019:GPTP}.

The initial population's genomes are generated using random sampling of a set of possible genes known as the genetic source~\cite{Helmuth:2020:ALife:source}. The sampling is weighted according to a target distribution of gene categories given in Table \ref{fig:gene-distribution}.

\begin{table}[t]
    \rowcolors{2}{gray!15}{white}
    \centering
    \begin{tabular}{l r}
        \toprule
        \textbf{Gene Category} & \textbf{Proportion} \\
        \midrule
        Variable & 0.2 \\
        Local Variable & 0.15 \\
        Literal & 0.15 \\
        Literal (via ERC) & 0.1 \\
        Abstraction & 0.15 \\
        Application & 0.15 \\
        Open \& Close & 0.1 \\
        \bottomrule
    \end{tabular}
    \caption{The distribution of gene categories for randomly generated genes and genomes, which is used to create the initial population and supply genes for mutation operators. Some literal genes are sampled from a discrete set dependent on the problem, while others were created via ephemeral random constant (ERC) generators~\cite{Poli:2008:field-guide-to-gp}. This distribution was selected on the basis of human intuition and is potentially sub-optimal.}
    \label{fig:gene-distribution}
\end{table}

\subsection{Compilation to AST}
\label{sec:compilation}

Linear genomes are translated into nested sequences prior to compilation. These nested sequences are referred to as \texttt{push} sequences because they are compiled using stacks by a process similar to the program execution model of PushGP systems. There are 2 stacks involved in compilation, one for typed ASTs and one for \texttt{push} sub-sequences called ``chunks''.

In addition to a \texttt{push} sequence, the compilation algorithm requires the type signature of the target program and a type environment containing the data types of each non-local variable. The type signature has 3 components: the return type, argument names, and argument types of the target program. Argument names are stored in an indexed collection of ``local'' variable names. Argument types are implicitly added to the type environment.

The compilation algorithm considers the elements of the \texttt{push} sequence one at a time and manipulates the stacks according to the kind of element.
When literal or variable expressions are encountered, they are pushed to the AST stack because they are valid, single-node, ASTs. Local variable genes, represented as an index, are resolved to a valid variable by looking up the variable name at position $mod(i,|locals|)$ where $i$ is the index provided by the gene and $locals$ is the set of local variables names containing, at least, the program's inputs. This is done because the full set of local variables that are ``in-scope'' at any particular point of compilation is not known ahead of time. The resulting variable expression is pushed to the AST stack.

When an function application instruction, \texttt{APP}, is encountered the AST stack is searched from top to bottom to find an AST with a function type. This could be a variable that corresponds to a function (such as $length$, $+$, and $map$) or an AST corresponding to a function abstraction expression. If no function type AST is found, the \texttt{APP} instruction is a noop. The compilation algorithm will then search for a collection of ASTs from the stack to use as arguments to the function AST. These argument ASTs must return a type compatible with the function's argument types. The search algorithm in Figure \ref{fig:compile-app-args} is used to find argument ASTs. If the algorithm fails to find an AST for any argument, the \texttt{APP} instruction is a noop. Otherwise, the function AST and all argument ASTs are removed from the AST stack and a new composite AST with an \texttt{APP} root node is pushed on top of the AST stack.

\begin{figure}
    \centering
    \begin{algorithmic}[1]
        \Require{$asts \gets $\texttt{the AST stack}}
        \Require{$f \gets$ \texttt{the function AST}}
        \State $argExprs \gets []$
        \State $substitutions \gets \{\}$
        \ForAll{$T \gets f.argTypes$}
            \State $T_2 \gets Substitute(T,\ substitutions)$
            \State $e \gets FindUnifiable(RemoveAll(asts,\ argExprs),\ T_2)$
            \State $argExprs \gets Append(argExprs,\ e)$
            \If{$e \not= NIL$}
                \State $newSubs \gets Unify(T_2,\ e.type)$
                \State $substitutions \gets Merge(substitutions,\ newSubs)$
            \EndIf
        \EndFor
        \State \textbf{return} $argExprs$
    \end{algorithmic}
    \caption{The algorithm for finding ASTs which will be type-safe arguments to the function being applied by the \texttt{APP} stack instruction. The result is a list of ASTs with $NIL$ values in the positions where no AST of a compatible type was found on the stack.}
    \label{fig:compile-app-args}
\end{figure}

If the function type of the function AST selected by the \texttt{APP} instruction is polymorphic (has a type scheme with a type variable), the call to the unification algorithm (Figure \ref{fig:compile-app-args}, line 8) will bind the free type variables to the concrete type of the argument AST. These bindings are substituted in later argument types (Figure \ref{fig:compile-app-args}, line 4) to ensure that all instances of the type variable are bound to the same type. For example, when compiling an \texttt{APP} with the function $append$ with type $\forall \alpha . (Sequence[\alpha], \alpha) \rightarrow Sequence[\alpha]$, the first argument can be satisfied by any sequence type. If the top-most AST with a sequence type has the type $Sequence[Int]$ then $\alpha$ will be bound to $Int$ which will become the second argument type of this particular application of $append$.

When a nested \texttt{push} sequence is encountered, it is pushed to the \texttt{chunk} stack. Function abstraction instructions and \texttt{LET} instructions use these chunks to compile a body of code that may reference the local variables created by those constructs, as described below.

When a function abstraction instruction (ie. \texttt{ABS[Int]}) is encountered new local variables are created to serve as the function's argument variables. No elements of the AST stack contain references to these new arguments names. The body of the function must be compiled from a \texttt{push} sequence within a new scope, or environment, that includes the new argument variables. To compile the body of the new function, the compilation algorithm makes a recursive call to itself with a new environment based on the \texttt{ABS} instruction's argument types and passes a \texttt{push} sequence from the \texttt{chunk} stack. During this nested call, local variable genes may resolve to one of the new local variables included in the environment. If the recursive compilation call does not produce any ASTs, for example if the chunk is empty or does not contain any leaf nodes from which to build ASTs, subsequent items of the \texttt{chunk} stack are compiled until an AST is found or no chunks remain on the stack. The latter scenario triggers the \texttt{ABS} instruction to noop.  Figure \ref{fig:compile-abs} details the process of compiling a function abstraction expression in pseudocode. 

\begin{figure}
    \centering
    \begin{algorithmic}[1]
        \Require{$abs \gets$ \texttt{the ABS instruction}}
        \Require{$asts \gets$ \texttt{the AST stack}}
        \Require{$chunks \gets$ \texttt{the Push chunk stack}}
        \Require{$locals \gets$ \texttt{the list of local variables}}
        \Require{$\Gamma \gets$ \texttt{the type environment}}
        \State $locals' \gets locals$
        \State $\Gamma' \gets \Gamma$
        \State $argVars \gets []$
        \ForAll {$argType \gets abs.argTypes$}
            \State $argVar \gets GenFreshVariable()$
            \State $argVars \gets Append(argVars, argVar)$
            \State $locals' \gets Append(locals',\ argVar)$
            \State $\Gamma' \gets Add(\Gamma',\ [argVar := argType])$
        \EndFor
        \ForAll {$chunk \gets chunks$}
            \State $funcBody \gets Compile(chunk,\ locals',\ \Gamma')$
            \If {$funcBody \not= NIL$}
                \State \textbf{return} $Fn(argVars, funcBody))$
            \EndIf
        \EndFor
        \State \textbf{return} $NIL$
    \end{algorithmic}
    \caption{The algorithm for compiling a function abstraction stack instruction. The result is an AST which with a \texttt{ABS} expression as the root. The type of AST is a function type with the argument types from the stack instruction and a return type determined by the type of the ``body'' expression compiled from the chunk.}
    \label{fig:compile-abs}
\end{figure}

When a \texttt{let} instruction is encountered, it is compiled similarly to a function abstraction instruction. A new local variable is created and the AST on top of the stack is used as its definition, which also determines its type. A \texttt{push} sequence from the \texttt{chunk} stack is then compiled with the new variable added to the set of local variables and the type environment. The AST resulting from the chunk compilation is used as the body of the \texttt{let} expression.

After all elements of the \texttt{push} sequence have been compiled, the AST stack will be a collection of type-safe ASTs that can be evaluated as programs. To select the single AST associated with the compiled genome, the top-most AST with a data type matching the target program's return type is selected.

Figure \ref{fig:neg-to-zero-genome} shows one possible genome that compiles to the solution of the negative-to-zero problem shown in Figure \ref{fig:neg-to-zero}. The slice of genome between the first pair of \texttt{OPEN} and \texttt{CLOSE} genes will be saved as chunk that will compile into a function body when the \texttt{ABS[Int]} stack instruction is processed. The first \texttt{LocalVar(1)} gene will resolve to the argument of the anonymous function created by \texttt{ABS[Int]}. The AST created by \texttt{ABS[Int]} will become the definition of the local variable produced by the \texttt{LET} at the end of the genome. The slice of genome between the second pair of \texttt{OPEN} and \texttt{CLOSE} genes will be saved as a chunk that is compiled into the body of the \texttt{LET} expression, and the second \texttt{LocalVar(1)} will resolve to the local variable defined by the \texttt{LET}. Notice that the \texttt{LET} expression is the root node of the AST (and the root symbol of the Clojure code in Figure \ref{fig:neg-to-zero-code}), therefore it is the last gene in the genome.

The genome in Figure \ref{fig:neg-to-zero-genome} is artificially simple for demonstration purposes. During evolution, most genomes are much longer and contain genes that either noop or build additional ASTs which do not get selected as the program because they do not return the correct data type for the problem or are buried deep in the stack at the end of compilation.

\subsection{Evolution}

For this work, a standard generational genetic algorithm was used to evolve programs. An initial population of random genomes was produced using the method described in Section \ref{sec:genomes}. Genomes are compiled into a type-safe ASTs (Section \ref{sec:compilation}) which are executed identically to a native function in the host language. These programs are evaluated based on a set of training cases in the form of input-output pairs.

The program's error on each training case is determined by a user provided error function. The collection of errors across all training cases is referred to as an individual's ``error vector''. If no AST with the problem's target return type is produced after compilation the individual is given a penalty error on every training case. If the compiled AST produces a runtime error, such as ``index out of bounds,'' when called on a training case, it is given a penalty error.

Parents are selected from the population of evaluated individuals on the basis of error vectors using Lexicase Selection~\cite{Helmuth:2015:UncompromisingLexicase, Helmuth:2019:LexicaseSpecialists}. The next generation of genomes is produced through variation of parent genomes.

If a individual is found to have an error of zero on all training cases, or if the maximum number of generations is reached, evolution is stopped and the individual with the lowest total error, given by the sum of its error vector, is returned. If this individual has a total error of zero, it is called a ``solution.''

\subsection{Simplification}

The best individual found during evolution is extracted for simplification. It has been shown that simplification acts as a form of regularization which improves the program's generalizability to unseen data cases~\cite{Helmuth:2017:simplification}. In addition, a simplified program may be easier for a human to understand. The best individual from evolution is simplified using a hill-climbing algorithm, as follows: 
\begin{enumerate}
    \item Create a new genome using an order-preserving random subset of the best individual's genome.
    \item Compile and evaluate the new genome to create a new individual. 
    \item If the total error of the new individual is equal to, or lower than, the current best individual it replaces the best individual.
    \item If iteration limit is reached, return best individual. Otherwise, return to step 1.
\end{enumerate}
The best individual's program after all iterations of simplification is reported as the output of the evolutionary search. For our experiments, this is the program that is tested for generalization on an unseen set of test cases.

\section{Experimental Design}
\label{sec:experiment}

We assess the ability of CBGP to perform automatic program synthesis using a subset of 14 problems from the program synthesis benchmark suite PSB1~\cite{Helmuth:2015:BenchmarkSuite}. The problems in PSB1 originate from introductory computer science textbooks, allowing us to assess how the system performs on the types of programming problems we ask new programmers to solve. We chose 14 problems that represent a wide range of requirements, such as data types and control flow, and difficulties. We purposefully avoided some problems that have typically been most difficult for other GP techniques and we have not yet benchmarked CBGP using the more recent (and difficult) problems of the suite PSB2~\cite{Helmuth:2021:GECCO:PSB2}, since this is CBGP's first benchmarking (and easier problems seem warranted) and because there is more data available for comparing CBGP with other program synthesis systems. However, assessing performance on PSB2 soon would supplement our experiments here.

Each problem is specified by a set of input/output examples defining the desired behavior of the program, in the form of supervised learning. Each run uses 100 training cases composed of hand-coded examples and a subset of a large set of randomly-generated inputs. Additionally, we use a set of 300 additional random examples to test each program that passes the training set for generalization; only those programs that perfectly pass all 300 examples are reported as solutions. The error functions used to measure the differences between program outputs and correct outputs are the same ones described in PSB1~\cite{Helmuth:2015:BenchmarkSuite}.

\begin{table}[t]
    \rowcolors{2}{gray!15}{white}
    \centering
    \begin{tabular}{l l}
      \toprule
      \textbf{Hyperparameter} & \textbf{Value} \\ \midrule
       Population Size & 1000 \\
       Max Generations & 300 \\
       Parent Selection & Lexicase Selection~\cite{Helmuth:2015:UncompromisingLexicase} \\
       Variation & UMAD~\cite{Helmuth:2018:GECCO:UMAD} \\
       Mutation Rate & 0.1 \\
       Simplification Steps & 2000 \\
       Initial Genome Sizes & [50, 250] \\
       Number of Training Cases & 100 \\
       Number of Unseen Test Cases & 300 \\
       \bottomrule
    \end{tabular}
    \caption{The evolutionary hyperparameters used for all runs of CBGP associated with the results presented in this paper.}
    \label{fig:hyperparameters}
\end{table}

We conduct 100 runs of CBGP per problem, and primarily report the success rate for each problem as measured by generalizing solutions. 
The generational genetic algorithm was configured with the hyperparameters given in Table \ref{fig:hyperparameters}. This configuration was selected due to its similarity to the configuration of PushGP in published results on the same set of benchmark problems. We leave the optimization of this configuration to future research.

With an aim to enhancing comparability between systems, we created a genetic source (function set) that largely matches that used in the comparison PushGP runs~\cite{Helmuth:2015:BenchmarkSuite, Helmuth:2018:GECCO:UMAD}. However, due to representational differences, there is not a one-to-one match between the genetic sources. The functions handling data operations are largely the same, but control flow is handled quite differently in CBGP with a functional host language compared to Push or the grammar-based programs of G3P and GE. In particular, much of the control flow in this implementation of CBGP is handled by higher-order functions that iterate over lists, such as \texttt{map}, \texttt{filter}, and \texttt{reduce}.

The description of the problems in PSB1 recommends not using every single available function for every problem~\cite{Helmuth:2015:BenchmarkSuite, Helmuth:2020:ALife:source}. For example, including functions that manipulate strings when solving a problem that only relates to lists of integers would expand the search space unnecessarily. As such, we follow these recommendations by creating type-tuned genetic sources for each problem in the fashion recommended by PSB1: for each problem, only include functions that manipulate the data types deemed relevant by PSB1. This ensures that we do not cherry-pick instructions known to be useful for a problem, while not including instructions that have no bearing on it.

\subsection{Comparison Methods}

We compare our CBGP results with those of other GP representations: PushGP, G3P, and GE. We choose comparison results from papers using comparable evolutionary hyperparameters as much as possible.

For PushGP, we use results from the paper introducing Uniform Mutation by Additions and Deletions (UMAD)~\cite{Helmuth:2018:GECCO:UMAD}. Like this paper, our CBGP runs use UMAD as the only genetic operator, making a reasonable comparison.

We use the paper introducing grammar design patterns as the results for G3P~\cite{Forstenlechner:2017:G3P}. The paper uses similarly type-tuned grammars to determine the instructions available to evolving programs.

Our reported GE results are taken from a paper exploring the use of domain knowledge and novelty in program synthesis~\cite{Hemberg:2019:DomainKnowledgeAndNoveltyImproveGE}. Since neither of those ideas are used in our work here, we use the baseline control results reported in the paper.

\section{Results}
\label{sec:results}

\begin{table*}[t]
\centering
\rowcolors{3}{gray!15}{white}
\begin{tabular}{l rrrr r}
\toprule
                           & \multicolumn{4}{l}{\textbf{Generalized Solution Rate}} & \multicolumn{1}{c}{\textbf{Generalization}} \\
\textbf{Problem}           & \multicolumn{1}{l}{\textbf{CBGP}}                  & \multicolumn{1}{l}{\textbf{PushGP}} & \multicolumn{1}{l}{\textbf{G3P}} & \multicolumn{1}{l}{\textbf{GE}} & \multicolumn{1}{c}{\textbf{Rate (CGBP)}} \\
\midrule
smallest                   & 100                                                & 100                                 & {\ul 89}                         & 100                              & 1.0                                              \\
mirror-image               & 100                                                & 100                                 & {\ul 1}                          & {\ul 25}                         & 1.0                                              \\
number-io                  & 100                                                & 98                                  & 96                               & 100                              & 1.0                                              \\
vectors-summed             & 100                                                & {\ul 11}                            & {\ul 85}                         & {\ul 1}                          & 1.0                                              \\
negative-to-zero           & 99                                                 & {\ul 80}                            & 98                               & {\ul 32}                         & 1.0                                              \\
median                     & 98                                                 & {\ul 55}                            & {\ul 65}                         & 99                               & 1.0                                              \\
vector-average             & 88                                                 & 88                                  & {\ul 0}                          & {\ul 0}                          & 0.99                                               \\
compare-string-lengths     & 22                                                 & 32                                  & {\ul 3}                          & 30                               & 0.79                                               \\
last-index-of-zero         & 10                                                 & \textbf{62}                         & \textbf{24}                      & 13                               & 0.92                                               \\
replace-space-with-newline & 0                                                  & \textbf{87}                         & 0                                & \multicolumn{1}{r}{-}           & \multicolumn{1}{r}{-}                            \\
small-or-large             & 0                                                  & \textbf{7}                          & 5                                & 0                                & \multicolumn{1}{r}{-}                            \\
count-odds                 & 0                                                  & \textbf{8}                          & \textbf{10}                      & 0                                & \multicolumn{1}{r}{-}                            \\ 
digits  & 0 & \textbf{19} & 0 & \textbf{70} & - \\ 
for-loop-index  & 0 & 2 & \textbf{25} & 0 & - \\ 
\bottomrule
\end{tabular}
\caption{Percentage of runs that found a generalized solution on each problem. Underlined values indicate the comparison method has a statistically significantly worse solution rate than CBGP according to a chi-squared test with at a p-value of 0.05. Values in bold indicate a statistically significantly better success rate using the same test. The generalization rate column denotes the proportion of runs for which the program which solved all training cases also solved the unseen test data.}
\label{table:solution_rates}
\end{table*}

Table~\ref{table:solution_rates} compares the success rates of CBGP to other GP representations on the 14 benchmark problems. CBGP performs quite well on 7 of the problems, producing success rates near or at 100. On all of these problems except number-io, at least one of the other methods performs significantly worse than CBGP. On the other hand, CBGP performs significantly worse than at least one other method on the last 6 problems. 

The last column in Table~\ref{table:solution_rates} gives the proportion of solutions on the training data that perfectly generalize to the unseen test set. Compared to the other three GP representations, which have typically produced low generalization rates on some, but not all, of these problems, the generalization rate of CBGP solutions is quite high across the board. For example, compare-string-lengths, last-index-of-zero, median, and negative-to-zero all produced generalization rates lower than 0.75, while almost no problem exhibited a generalization rate of 1.0, in a study of generalization using PushGP~\cite{Helmuth:2017:simplification}.

\begin{table}[t]
\centering
\caption{Solution sizes for each problem that CBGP solved. Min gives the minimum size of any solution program, while Pre and Post give the mean sizes before and after applying automatic simplification to the solution genomes.}
\label{table:sizes}
\rowcolors{2}{gray!15}{white}
\begin{tabular}{lrrr}
\toprule
\textbf{Problem}                & \textbf{Min} & \textbf{Pre}   & \textbf{Post}  \\
\midrule
smallest               & 7   & 7.55  & 7.18  \\
mirror-image           & 4   & 4.54  & 4.06  \\
number-io              & 4   & 4.92  & 4.03  \\
vectors-summed         & 4   & 4.25  & 4.00     \\
negative-to-zero       & 7   & 7.90  & 7.02  \\
median                 & 9   & 10.52 & 10.03 \\
vector-average         & 7   & 9.74  & 8.89  \\
compare-string-lengths & 10  & 12.34 & 11.79 \\
last-index-of-zero     & 8   & 12.42 & 10.33 \\
\bottomrule
\end{tabular}
\end{table}

Table~\ref{table:sizes} presents the sizes of solution programs found for each problem solved by CBGP. Program size is measured in number of nodes in the Clojure S-expression representation of the program, which is identical to the number of nodes in the AST. We find that three of these problems have been solved by programs containing only 4 nodes, while the remainder have been solved by programs with 10 or fewer nodes. Interestingly, the mean solution sizes pre- and post-simplification tend to be quite close to the minimum sizes. This means that evolved solutions rarely have unnecessary code in the programs themselves. Note that genomes, on the other hand, may have lots of unnecessary genes that either produce unused ASTs or have no effect on AST compilation. It seems that removing such unnecessary genes during simplification does not result in dramatically simpler programs, as the mean post-simplification program size is not much smaller than pre-simplification.

\subsection{Example Solution Programs}

The supplementary materials to this paper include a file containing every solution evolved by CBGP. In Figure \ref{fig:example-solutions} we give some examples of those solution programs and note some of their interesting features below.

The solution to negative-to-zero interestingly maps the subtract function over two copies of the input vector, which produces a vector entirely made of zeros. It then maps the \texttt{max} function over the zeros and the input vector, changing every negative integer into 0 as required.

The vector-average solution behaves as expected. One thing to note is that it converts the length of the input vector to a $Double$, since the \texttt{count} function is typed to return an $Int$. Future work into allowing for subtyping or type classes could allow for all $Int$ expressions to be considered valid $Double$ expressions, but for now, the conversion must happen explicitly.

The smallest problem requires the program to find the minimum of four inputs. Instead of simply applying the \texttt{min} function 3 times, this solution unnecessarily defines a new function that finds the \texttt{min} of \texttt{input4} and its argument, and then applies that function to \texttt{input1}.

The last-index-of-zero problem requires the program to find the last index where 0 appears in the input list. This solution reverses the input, finds the first index of zero, and then subtracts that from the decremented length of the input. This strategy is similar to some solutions to this problem that have been evolved in PushGP.

\begin{figure}
\begin{verbatim}
(defn negative-to-zero
  [input1]
  (map max (map - input1 input1) input1))

(defn vector-average
  [input1]
  (safe-div (reduce + input1)
            (float (count input1))))

(defn smallest
  [input1 input2 input3 input4]
  (min (min input3 
            ((fn [a-639347] (min input4 a-639347)) input1))
       input2))

(defn last-index-of-zero
  [input1]
  (- (count (butlast input1))
     (index-of (reverse input1) 0)))
\end{verbatim}
\caption{A sample of solution Clojure programs evolved by CBGP. Anonymous function argument symbols were generated using a unique natural number prefixed with an \texttt{a-}. Whitespace was adjusted for readability.}
\label{fig:example-solutions}
\end{figure}


\section{Discussion and Future Work}

CBGP has demonstrated that it can readily find solutions to some problems, but on others the solution rate of CBGP quickly drops to zero. These trends correlate somewhat to the problems found difficult by other GP representations; however, there are some problems that CBGP solves readily that others do not and vice versa.
When initially tested on the PSB1 benchmark problems, the other genetic programming systems saw similar trends, and have since increased performance as the methods mature through continued research. We hope to see a similar rise in the search performance of CBGP in the future.

The large variety in which problems each system finds easier or harder points to the importance of program representation for search performance. This area is not well understood in GP, and we hope that CBGP can help better illuminate important differences in representation. We suggest this as an area of future research such that we can understand what makes problems difficult under a given representation.

One hypothesis regarding different representations producing wildly different results on some problems is the impact of representations on the size of programs needed to form a particular computation. Solutions to problems with high solution rates tend to be smaller than solutions to problems with a low success rate, regardless of representation. The problems that CBGP solves most readily have small solution programs, and similar results have been shown for PushGP on the same problems~\cite{Helmuth:2015:BenchmarkSuite}. We do not know if similarly small solution programs are possible for the problems that CBGP did not solve, and it simply did not find them, or if they require larger programs and therefore are more difficult to find in the search space. Further research into CBGP solutions to these problems could help us understand whether it is simply the size of the solution programs preventing them from being solved, or whether CBGP has issues traversing the search space effectively regardless of solution size for some problems.

The exceptionally high generalization rate of CBGP is not easily explained. When considered in combination with the inability to find solutions on harder problems, this may be an indication that CBGP cannot fall back on memorizing or bloating the program into something that overfits the training data. In CBGP, an increase in genome size does not necessarily cause an increase in program size because additional genes may simply result in more ASTs being left on the stack after compilation, rather than larger ASTs. This hypothesis is further supported by the minimum and average program sizes of solutions found by CBG. The problems with high solution (and generalization) rates are solved by small programs. 

When looking through solution programs, we found very few instances of programs that define and use anonymous functions effectively. Defining such functions is an integral part of functional programming for human programmers. Thus one piece of important future work is to try to assess why CBGP is not making use of function definition, and considering ways to encourage this behavior.

One insight from this research that may be helpful to the wider research fields of genetic programming and program synthesis is the value of introducing formalisms, such as type theory, into our systems. The body of work accumulated in fields of theoretical computation provide the program synthesis community with tools to guide synthesis towards programs with desirable properties, such as type safety.

Functional CBGP can, in theory, represent programs using any data type or language construct supported by the type system, and the unification algorithm in particular. A valuable direction of future research is to implement the common extensions to the Hindley-Milner type system which add support for function overloading, sub-types, and variadic functions~\cite{Smith:1992:TypeInferenceOverloadingSubtyping, Pottier:1998:TypeInferenceWithSubtyping, CARDELLI:1994:SystemFSubtyping, Dolan:2017:PolymorphismSubtypingTypeInference}. The primary benefit of these extensions would be the ability to represent programs using all the features of a modern functional programming language and possibly approach any programming task that can be well-specified by types. Another benefit of supporting additional kinds of polymorphism is the ability to use a smaller genetic source with considerably fewer, more general, functions which could dramatically reduce the search space and improve solution rates on complex problems.

\section{Conclusion}

In this paper we present functional Code Building Genetic Programming and show how it leverages type theory to ensure synthesized programs are type safe while also allowing polymorphic functions, anonymous functions, and higher order functions to be expressed. We report on empirical benchmarks that show CBGP can find solution programs more consistently than other contemporary GP methods on some problem, while it struggles to find any solutions on others. Investigations into solution programs show repeated use of polymorphic functions and higher order functions, but little use of anonymous function definitions.

When CBGP does find a solution on training data, we observe an exceedingly high rate of generalization to unseen test data. This phenomenon is in contrast to the comparatively low generalization rates of all other GP systems included in our comparison~\cite{Helmuth:2017:simplification,Forstenlechner:2017:G3P,Sobania:2021:EuroGP}. Furthermore, the solution programs found by CBGP are small, even without the use of typical genome simplification techniques.

Finally, we direct future research towards a deeper utilization of type theory in general program synthesis systems. We also suggest the genetic programming field perform broader studies into the impact of representation on problem difficulty.

\begin{acks}
This material is based upon work supported by the National Science Foundation under Grant No. 1617087. Any opinions, findings, and conclusions or recommendations expressed in this publication are those of the authors and do not necessarily reflect the views of the National Science Foundation.
\end{acks}

\bibliographystyle{ACM-Reference-Format}
\bibliography{fcbgp-bib} 

\end{document}